\documentclass[conference]{IEEEtran}
\IEEEoverridecommandlockouts
\usepackage{cite}
\usepackage{amsmath,amssymb,amsfonts}
\usepackage{graphicx}
\usepackage{textcomp}
\usepackage{xcolor}
\usepackage{color,soul}
\usepackage[ruled,linesnumbered]{algorithm2e}

\def\BibTeX{{\rm B\kern-.05em{\sc i\kern-.025em b}\kern-.08em
    T\kern-.1667em\lower.7ex\hbox{E}\kern-.125emX}}
\begin{document}

\title{A Comprehensive Analysis on Adversarial Robustness of Spiking Neural Networks\\
\thanks{This work was supported in part by C-BRIC, one of six centers in JUMP, a Semiconductor Research Corporation (SRC) program sponsored by DARPA.
}
}

\author{\IEEEauthorblockN{Saima Sharmin*, Priyadarshini Panda*, Syed Shakib Sarwar, Chankyu Lee,  Wachirawit Ponghiran  and Kaushik Roy}
\IEEEauthorblockA{\textit{School of Electrical and Computer Engineering,}
\textit{Purdue University}\\ West Lafayette, IN 47907, USA \\
email: \{ssharmin, pandap, sarwar, chankyu, wponghir,  kaushik\}@purdue.edu
}
* Equal author contribution \\

}

\maketitle

\begin{abstract}
In this era of machine learning models, their functionality is being threatened by adversarial attacks. In the face of this struggle for making artificial neural networks robust, finding a model, resilient to these attacks, is very important. In this work, we present, for the first time, a comprehensive analysis of the behavior of more bio-plausible networks, namely Spiking Neural Network (SNN) under state-of-the-art adversarial tests. We perform a comparative study of the accuracy degradation between conventional VGG-9 Artificial Neural Network (ANN) and equivalent spiking network with CIFAR-10 dataset in both whitebox and blackbox setting for different types of single-step and multi-step FGSM (Fast Gradient Sign Method) attacks. We demonstrate that SNNs tend to show more resiliency compared to ANN under blackbox attack scenario. Additionally, we find that SNN robustness is largely dependent on the corresponding training mechanism. We observe that SNNs trained by spike-based backpropagation are more adversarially robust than the ones obtained by ANN-to-SNN conversion rules in several whitebox and blackbox scenarios. Finally, we also propose a simple, yet, effective framework for crafting adversarial attacks from SNNs. Our results suggest that attacks crafted from SNNs following our proposed method are much stronger than those crafted from ANNs.
\end{abstract}


\section{Introduction}
Contemporary machine learning models like Artificial Neural Networks (ANN) have completed several milestones towards gaining super-human performance in visual recognition tasks like image classification, text and voice recognition \cite{b1},\cite{hinton} etc. Application of such networks is being considered for autonomous cars, drones and robotics. For such mission critical applications, there is an urgent need to improve the robustness of networks against adversarial attacks.
Adversarial attacks \cite{szegedy},\cite{goodfellow} can be generated by injecting carefully-crafted perturbations to a clean input so that it can deceive the model into producing incorrect outputs with high probability. Note, the perturbation is small enough to be perceptible to the human eye. This vulnerability holds even when the adversarial input is generated from a different trained model other than the target \cite{papernot}. The profound implication of the problem has triggered research interest towards addressing this issue and finding ways to defend against adversarial attacks in the context of state-of-the-art neural network models.

Research in \cite{goodfellow, szegedy, sharif, kurakin2} shows the vulnerability of deep ANNs against adversaries. Several adversarial training and defense mechanisms have been proposed in this regard, like ensemble training \cite{ensemble}, implicit prior modeling with random noise \cite{exL}, scalable training \cite{kurakin} etc. However, one fundamental question that remains unanswered is whether there are any network models inherently resistant to adversarial attacks. In the face of this question, more biologically plausible neural network model, like Spiking Neural Network (SNN) comes into picture. In an SNN, the network receives stochastic stimulation from noisy neurons in the form of Poisson spike train, leading to the temporal evolution of the membrane potential of the neurons \cite{snn}. This inherent noise embedded in an SNN makes it worthwhile to investigate how the spiking network reacts under adversarial attacks, compared to ANNs.

In this paper, we have analyzed the behavior of large-scale spiking neural networks against state-of-the-art adversarial attacks. Our experiment is focused on VGG-9 network models with CIFAR-10 dataset. To the best of our knowledge, this is the first work to analyse the characteristics of a spiking network under different kinds of adversarial attacks. The key contributions can be summarized as follows:  
\begin{itemize}
\item Spiking Neural Network model is a comparatively newer addition to the machine learning family. Although there exists significant amount of literature on the technique of crafting adversarial input from ANN, there is none to generate SNN-crafted adversary. \textit{We propose a simple mechanism to generate adversarial inputs from SNN model parameters without the need of any non-trivial gradient calculation in the spiking domain.}
\item We present comprehensive quantitative comparison of the behavior of ANN and equivalent SNN under different attack scenarios. \textit{We observe that spiking networks are more robust than rate-based ANNs for blackbox attacks (i.e. when attacker has no knowledge of the target model's parameters). In whitebox attack (i.e. when attacker has full knowledge of the target model's parameters), SNNs, generally, yield higher accuracy degradation than ANNs. Furthermore, our results suggest that attacks crafted from SNNs following our proposed method are much stronger than those crafted from ANNs.} 
\item We demonstrate that the adversarial resistance of SNNs varies depending on the training mechanism. We have considered two different training methods: ANN-to-SNN conversion\cite{diehl,abhronil} and direct spike-based backpropagation\cite{direct_train1, direct_train} in our experiments. \textit{We observe that the latter method shows better resistivity under whitebox and blackbox scenarios.}
\end{itemize}

The organization of the rest of the paper starts with illustration of the basic concepts of the adversarial attacks and Spiking Neural Network (SNN) in section \ref{sec:attack_basic} and \ref{sec:snn_basic}, respectively. In the next section (section \ref{sec:exp}), we explain the network architecture, training methods, adversarial input generation and testing process. Section \ref{sec:results} contains our simulation and analysis results, followed by conclusion in section \ref{sec:conclusion}. \\

\section{Adversarial attack: Fundamentals}\label{sec:attack_basic}
Given a classification model $h$ with dataset $(x, y_{true})$, where $x$ is the clean image and $y_{true}$ is the corresponding correct label, the main concept of adversarial attack is to find an input $x^{adv}$ such that $x$ and $x^{adv}$ are indistinguishable to the human eye, yet the model misclassifies $x^{adv}$, i.e. produces high probability on wrong labels. In our work, we have considered the following approaches to generate $x^{adv}$. \\
\subsection{Non-targeted FGSM (Fast Gradient Sign Method)} \label{sec:fgsm}
This is the most basic and widely used approach to generate adversarial perturbations in the following form \cite{goodfellow}
\begin{equation}
    \label{eqn:fgsm}
    x^{adv} = x + \epsilon sign({\nabla_x}J(x, y_{true}))
\end{equation}
Here $\epsilon$ refers to the amount of perturbation. Usually, the value of $\epsilon$ is much smaller than the unperturbed data $x$. $J(x,y_{true})$ is the loss function of the model. ${\nabla_x}J$ is the gradient of the loss function with respect to the original clean data. \\
\subsection{Non-targeted R-FGSM (Random-step FGSM)}\label{sec:rfgsm}
In this method, the single step gradient calculation is preceded by a simple step of adding small random noise ($N(O^d, I^d)$) to the image beforehand.
\begin{equation}
    \label{eqn:rfgsm1}
    x' = x + \alpha sign(N(O^d, I^d)),
\end{equation}
\begin{equation}
    \label{eqn:rfgsm2}
    x^{adv} = x' + (\epsilon-\alpha) sign({\nabla_{x'}}J(x', y_{true}))
\end{equation}
Here, initial perturbation $\alpha < \epsilon$. 
Authors in \cite{ensemble} introduced this method to escape the non-smooth vicinity of the data point. \\
\subsection{I-FGSM (Iterative FGSM)}\label{sec:ifgsm}
This is a multistep method for generating adversarial inputs. It iteratively applies FGSM with step-size $\alpha \geq \epsilon/k$, where $k$ denotes the number of iterations \cite{ensemble}. 
In non-targeted I-FGSM, the loss is calculated with respect to the true label $y_{true}$, whereas targeted I-FGSM uses either a random class, $y_{random}$ or the least likely class, $y_{ll}$ for calculating loss function and perturbs the input in the opposite direction as the gradient.\\

\subsubsection{Non-targeted I-FGSM}
\begin{equation}
    \label{eqn:ntifgsm1}
    {x_0}^{adv} = x,
\end{equation}
\begin{equation}
    \label{eqn:ntifgsm2}
    {x_{N+1}}^{adv} = Clip_{x,\epsilon}\{{x_{N}}^{adv} + \alpha sign({\nabla_x}J({x_{N}}^{adv}, y_{true}))\}
\end{equation}
\subsubsection{Targeted I-FGSM}
\begin{equation}
    \label{eqn:tifgsm1}
    {x_0}^{adv} = x,
\end{equation}
\begin{equation}
    \label{eqn:tifgsm2}
    {x_{N+1}}^{adv} = Clip_{x,\epsilon}\{{x_{N}}^{adv} - \alpha sign({\nabla_x}J({x_{N}}^{adv}, y_{ll}))\}
\end{equation}
${x_N}^{adv}$ is the adversarial sample at $N^{th}$ iteration. $y_{true}$ and $y_{ll}$ are the correct and the least-likely class label, respectively. $\alpha$ is the perturbation per step. $Clip_{x,\epsilon}()$ denotes element-wise clipping of the argument to the range $\lbrack x-\epsilon, x+\epsilon\rbrack$ \\

\section{Spiking Neural Network: Fundamentals}\label{sec:snn_basic}
Spiking Neural Networks (SNN) operate based on bio-plausible event-driven algorithm. From network topology perspective, the activation blocks (like Rectified Linear Unit) of the ANN is replaced by biological neuron-based functional blocks (e.g. Integrate and Fire (IF) neuron, Leaky Integrate and Fire (LIF) neuron) in the equivalent SNN. The dynamics of LIF spiking neuron is formulated as follows:
\begin{equation}
    \label{eqn:ntifgsm}
    \tau \frac{dV_{mem}}{dt} = -V_{mem}+w\theta(t-t_k)
\end{equation}
$V_{mem}$ is the membrane potential of the neurons, $\tau$ is the time constant for the decay of $V_{mem}$, $w$ is the synaptic weight and $\theta(t-t_k)$ represents the spike at time instant $t_k$.
There are mainly two broad categories for training an SNN: unsupervised and supervised. However, in this work, we have used two of the supervised training strategies in order to achieve high accuracy. A brief illustration of these two techniques are presented in the next two subsections.\\ 

\subsection{ANN to SNN conversion (SNN-I)}\label{sec:ann_to_snn}
ANN to SNN conversion method considers a simple Integrate and Fire neuron (IF) as the neuron activation function due to its functional resemblance to Rectified Linear Unit (ReLU), without any leak or refractory period. Several authors \cite{diehl, abhronil} have proposed techniques of adjusting the synaptic weights ('weight normalization') or neuronal threshold values ('threshold balancing') to acquire lossless transformation from ANN to SNN. The accuracy reported for SNNs, trained in this way, is high, compared to ANN, even for very large scale networks.\\

\subsection{Spike-based training (SNN-II)}\label{sec:spike_train}
In this method, SNN is directly trained based on an event-driven supervised gradient descent backpropagation algorithm. Unlike the conversion mechanism, LIF neurons are used as the basic building block here. In forward propagation, Poisson-distributed spike train, generated from the input pixels, are fed to the network. Accumulated weighted spikes at the input of a neuron, at time $t$, triggers an output spike, if it exceeds a threshold value. 
Neurons at each layer undergo this process based on the input spikes received from the preceding layer. In order to carry out backpropagation in the spiking domain, we need a differentiable transfer function for the neurons. To that effect, the activation of the spiking neuron is formulated by low-pass filtering the spike train, according to the following equation.
\begin{equation}
    \label{eqn:neuron_act}
    Activation, a(t) = \frac{1}{T} \sum_{k=1}^{t}exp(-\frac{t-t_k}{\tau})
\end{equation}
The time constant $\tau$ dictates the decay rate of the neuron activation. $T$ is the total time. $t_k$ refers to the time instant of the $k$-th spike.
During the backpropagation process, the gradient of error with respect to weight requires calculating the derivative of the neuron activation $a$ with respect to the net input to the neurons, which is approximated by the following equation:
\begin{equation}
    \label{eqn:diff_neuron_act}
    \frac{\delta a}{\delta net} = \frac{1}{V_{th}}(1+\frac{\delta a}{\delta t})
\end{equation}
where $V_{th}$ refers to the threshold value of the neuron, $net$ is the accumulated weighted sum of spikes at the input of a neuron, and $t$ is the time instant.
The details of the backpropagation algorithm is illustrated in \cite{direct_train1, direct_train}. \\

\section{Experiments}\label{sec:exp}
\subsection{Dataset and network topology}
Our experiments mainly focus on the standard visual recognition dataset CIFAR-10 with VGG-9 networks. The VGG-9 architecture is
32$\times$32-64c5-64c5-2s-128c5-128c5-2s-256c5-256c5-256c5-2s-1024fc-10o, where c = convolutional layer, s = sub-sampling layer, fc = fully connected layer and o = output layer. The input image in CIFAR-10 dataset results in 3-channel 32$\times$32 input neurons, followed by 2 subsequent layers of 64 convolutional kernels of size 5$\times$5 each, followed by 2$\times$2 spatial averaging sub-sampling window. This convolution process is repeated in the second and the third stage with 128 and 256 maps of convolutional kernels, respectively. Note that the third stage has 3 convolutional layers. The final two stages of the networks are fully connected layers. The outputs from the third stage sub-sampling is vectorized and fed into a fully connected layer with 1024 outputs. The final layer consists of 10 output neurons corresponding to the 10 classes of CIFAR-10. It is worth-mentioning that each of the convolutional, sub-sampling and fully connected layers are followed by LIF neurons (ReLU activations) in SNN (ANN) architecture. We have also used a dropout of 0.2 after each convolution and fully-connected layer.\\

\subsection{ANN Training}\label{sec:ann_train}
\begin{figure*}[!t]
\centerline{\includegraphics{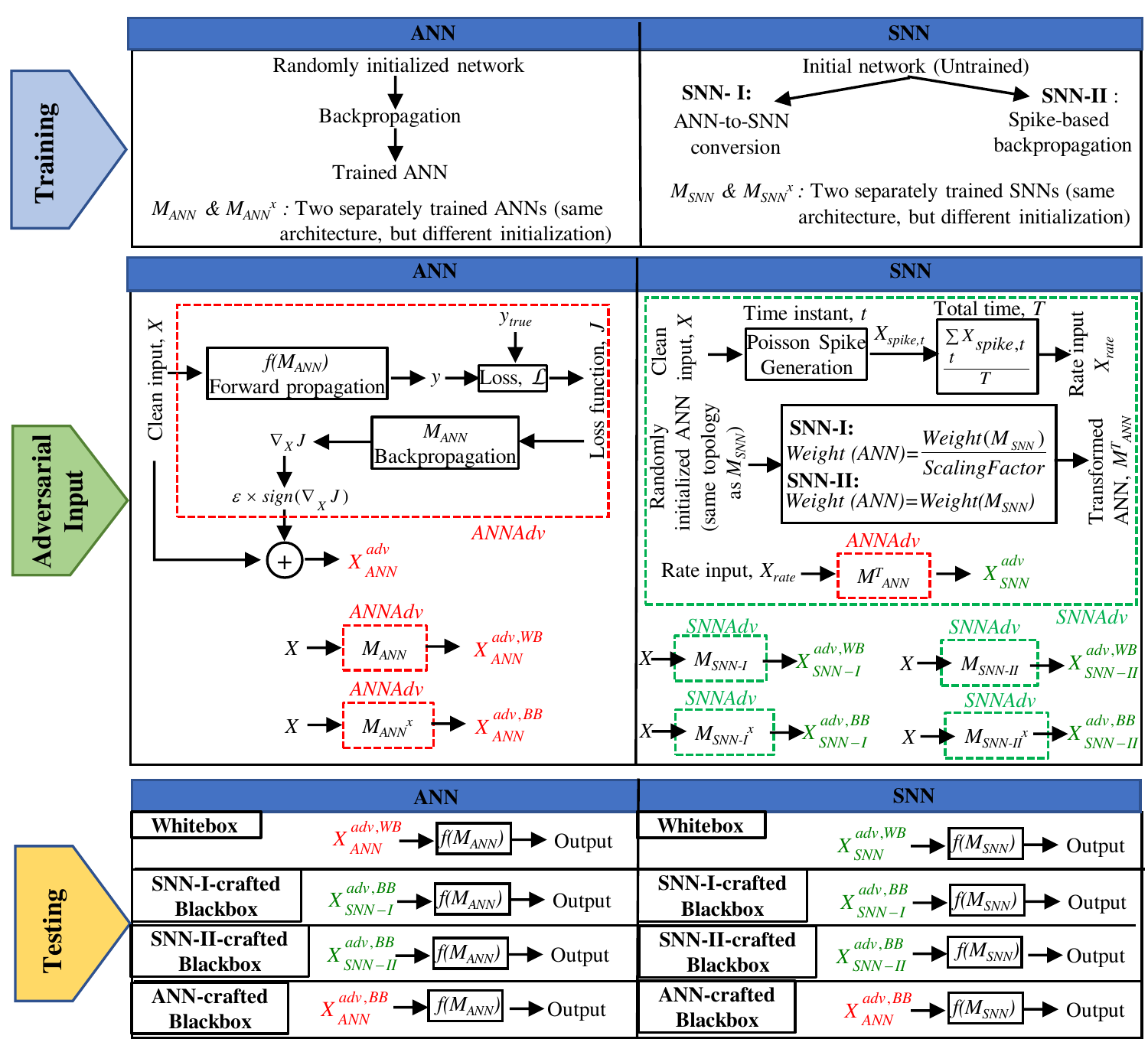}}
\caption{A schematic summarizing the training, adversarial input generation and testing methods used in our work for both Artificial and Spiking Neural Networks (ANN and SNN). \textbf{Training:} ANNs are trained according to the standard backpropagation formula, while two different techniques have been followed for training SNNs: ANN-to-SNN conversion\cite{diehl, abhronil} and direct-spike based backpropagation\cite{direct_train1, direct_train}. The generated SNNs are referred as SNN-I and SNN-II, respectively. \textbf{Adversarial input generation:} The flowchart inside the dotted red box, labelled as \textit{ANNAdv}, refers to the Fast Gradient Sign Method (FGSM)\cite{goodfellow}. According to this method, adversarial input for $M_{ANN}$ is produced from the clean input perturbed by the sign of the gradient of the loss function with respect to the input (${\nabla_X}J$). $\epsilon$ is the amount of perturbation. In case of generating adversary from SNNs, first, we have created a transformed ANN (${M^T}_{ANN}$) by loading the weights (or scaled weights, in case of SNN-I) of the trained SNNs into a randomly initialized ANN. Next, rate-encoded input $X_{rate}$ is passed through the \textit{ANNAdv} method, with ${M^T}_{ANN}$. Dotted green box is used to represent SNN-crafted adversary generation method, labelled as \textit{SNNAdv}. We have trained two separate networks with the same architecture but different initializations for each of ANN, SNN-I and SNN-II (denoted as $M_{ANN}$, $M_{{ANN}^x}$ for ANN; $M_{SNN-I}$, $M_{{SNN-I}^x}$ for SNN-I and $M_{SNN-II}$, $M_{{SNN-II}^x}$ for SNN-II). When $M_{ANN}$ is used as both the target and source model, the generated adversarial input is labelled as ${X_{ANN}}^{adv,WB}$, where the superscript WB stands for Whitebox. On the other hand, adversarial inputs crafted from $M_{{ANN}^x}$ (for attacking the target $M_{ANN}$) are called ${X_{ANN}}^{adv,BB}$ (BB for Blackbox). Same labelling strategy is used for SNN adversary generation. \textbf{Testing:} The target models used for test purpose are $M_{ANN}$, $M_{SNN-I}$ and $M_{SNN-II}$. We have performed four different comparisons between ANN and SNN adversarial attacks. Whitebox: ANN and SNN (both SNN-I and SNN-II) models are attacked by adversarial inputs crafted from the same model as the target.  SNN-I-crafted (SNN-II-crafted) Blackbox: ANN and SNN models are tested with a common adversary: SNN-I-crafted (SNN-II-crafted) blackbox samples ${X_{SNN-I}}^{adv,BB}$ (${X_{SNN-II}}^{adv,BB}$). ANN-crafted Blackbox: ANN and SNN models are tested with a common set of adversary: ANN-crafted blackbox samples ${X_{ANN}}^{adv,BB}$. Note, red inputs are used to indiacte ANN-crafted samples, while the green ones are SNN-crafted. The results of these four comparisons are presented in the four columns in Fig. \ref{fig:attack_types}.}
\label{fig:workflow}
\end{figure*}

The first step of our experiment consists of training the ANN models (network topology described in the previous section), as showed in Fig. \ref{fig:workflow}. Training of VGG-9 ANN is performed with 200 epochs at an initial learning rate of 0.09, which is reduced by a factor of 10 at the $81^{st}$ and the $122^{nd}$ epoch (also known as learning rate annealing) in order to ensure gradual decrease of the loss function during the training process. Our custom simulation framework is implemented based on PyTorch deep learning library\cite{pytorch}.

\subsection{SNN Training}\label{sec:snn_train}
In order to train the spiking VGG-9 model by following the conversion method\cite{abhronil} (SNN-I), we have adjusted layer-by-layer neuronal threshold values (theory in sec \ref{sec:ann_to_snn}) with the maximum membrane potential at the corresponding input, by running the forward propagation sequentially for each layer. We have used a total of 2000 time steps for the entire forward propagation, since it demands a sizable time-window to find the optimum threshold values.

On the other hand, in case of the spike-based backpropagation training of SNN (SNN-II), total number of time steps is 70 during the forward propagation. The training is performed with 125 epochs, where the learning rate is reduced at the $70^{th}$ and the $100^{th}$ epoch. The details of the training theory can be found in sec. \ref{sec:spike_train} and Ref. \cite{direct_train}.
The baseline accuracy of the ANN and the trained SNNs are summarized in table \ref{table:baseline}.\\

\begin{table}[!t]
\caption{Baseline accuracy}
\label{table:baseline}
\begin{center}
    \begin{tabular}{| c | c |}
    \hline
    Dataset  & CIFAR-10
    \\ \hline
    Network topology  & VGG-9
    \\ \hline
    ANN accuracy  & 89.5\%  
    \\ \hline
    SNN-I$^{\mathrm{a}}$ accuracy & 85.6\% 
    \\ \hline
    SNN-II$^{\mathrm{b}}$ accuracy & 87.1\% 
    \\ \hline
    \hline
    \multicolumn{1}{l}{$^{\mathrm{a}}$ANN-to-SNN conversion. $^{\mathrm{b}}$Spike-based backpropagation..}
 
    \end{tabular}
\end{center}
\end{table}

\subsection{Adversarial input generation: ANN-crafted}\label{sec:ann_crafted}
After the completion of the training phase, adversarial inputs are generated from the trained models using four different methods: (i) Non-targeted FGSM, (ii) Non-targeted R-FGSM, (iii) Non-targeted I-FGSM and (iv) Targeted I-FGSM. The flowchart in Fig. \ref{fig:workflow} elaborately describes the FGSM method of adversary generation. According to (\ref{eqn:fgsm}), we have calculated the sign matrix from the input gradient and multiplied with perturbation $\epsilon$. Since we normalized the image dataset to represent zero mean, the absolute values of the input pixel intensity in clean images range from 0 to 1. Hence, the amount of perturbation inflicted upon the pixels needs to be normalized too. We have used $\epsilon$ = 8/255, 16/255, 32/255 and 64/255 (in some cases) in our experiment. In the Iterative FGSM,  we have experimented with two-step, five-step and seven-step iterative methods in order to investigate the effect of the number of iteration steps. Note that we have used a random class as $y_{ll}$ in (\ref{eqn:tifgsm2}) for Targeted I-FGSM. All of the parameters used in (\ref{eqn:fgsm}) - (\ref{eqn:tifgsm2}) have been summarized in table \ref{table:attack_params}.
\begin{table}[!t]
\caption{Adversarial attack: parameters}
\label{table:attack_params}
\begin{center}
    \begin{tabular}{| c | c | c | c |}
    \hline
    \textbf{Attack type} & \textbf{Perturbation, $\epsilon$} & \textbf{no. of steps, $k$} & \textbf{$\alpha$} \\ \hline
    FGSM & 8/255, 16/255, 32/255 & 1 & -
    \\ \hline
    R-FGSM & 8/255, 16/255, 32/255 & 2 & $\epsilon/2$
    \\ \hline
    I-FGSM  & 8/255, 16/255, 32/255 & 2, 5, 7 & $\epsilon/k$
    \\ \hline
    \hline
    \end{tabular}
\end{center}
\end{table}

\SetKwInput{Input}{Input}
\SetKwInput{Output}{Output}
\begin{algorithm}
    \caption{\textbf{ANN-crafted input:} \textit{ANNAdv}($X$, ${NN}$, $\epsilon$)}
    \label{algo1}
    \SetAlgoLined
        \Input{Clean dataset $X$, True label $Y_{true}$, Trained ANN model ${NN}$ with parameter set $\theta$, forward propagation $f_{NN}$, adversarial perturbation $\epsilon$}
        \Output{Adversarial dataset $X_{adv}$}
        \For {minibatch $X^i$ in $\left \{X^1,...,X^m\right\}$} {
            \textbf{forward:}{Output $Y = f_{NN}(X;\theta)$ }\\
            \textbf{Loss:}{$\mathcal{L}(Y, Y_{true})$ }
            \textbf{Backward:}{${\nabla_\theta}\mathcal{L}$ } \\
            \textbf{Adversarial data:}{${X^i}_{adv} = X^i + \epsilon \times sign({\nabla_{X^i}}\mathcal{L})$ }}
\end{algorithm}

\begin{algorithm}
    \caption{\textbf{SNN-crafted input:} \textit{SNNAdv}($X$, ${SNN}$, $\epsilon$)}
    \label{algo2}
    \SetAlgoLined
        \Input{Clean dataset $X$, True label $Y_{true}$, Trained SNN model ${SNN}$ with parameter set $\theta$, forward propagation $f_{SNN}$, adversarial perturbation $\epsilon$}
        \Output{Adversarial dataset $X_{adv}$}
        {Random initialization of an equivalent ANN model $NN' (\theta, f_{NN'})$}\\
        {Modification of $NN'$ weights, $W_{NN'} = W_{trainedSNN}$} \\
        \For {minibatch $X^i$ in $\left \{X^1,...,X^m\right\}$} {
            \For{each timestep $t$ in total time $T$} {\textbf{Poisson Spike Train:} ${X_t}^i$}
            \textbf{Rate input:}{${X^i}_{rate} = \frac{\sum_t{{X_t}^i}}{T}$}\\
            \textbf{Adversarial data:}{${X^i}_{adv} = ANNAdv({X^i}_{rate}, NN', \epsilon)$ }}
\end{algorithm}

\subsection{Adversarial input generation: SNN-crafted}\label{sec:snn_crafted}
For a comprehensive analysis and comparison, we have devised a method to generate attack samples from SNNs as well. Algorithm 1 describes the widely-known FGSM in the context of ANN. Algorithm 2 illustrates its proposed SNN counterpart.
FGSM calculates the gradient of the loss function with respect to the clean input data. Due to the non-trivial operations during gradient calculation in a spike-based model, we have come up with a simple framework. Initially, an ANN model, $NN'$ with the same network topology is randomly initialized. The SNN model is independently trained ($M_{SNN}$). Subsequently, $NN'$ weight matrices are mapped and overwritten with the learned weights of $M_{SNN}$. Next, rate-based input, $X_{rate}$ is generated from the Poisson spike train of the clean dataset. Afterwards, $X_{rate}$ and $NN'$ model are used to generate FGSM adversarial input following Algorithm 1. This method has been schematically illustrated in Fig. \ref{fig:workflow}. Note, for SNN-I, the scaling factors of the weights of the transformed ANN, ${M^T}_{ANN}$ (Fig. \ref{fig:workflow}) equal the threshold-scaling factors used in the ANN-to-SNN conversion mechanism during training.

\subsection{Testing}\label{sec:test}
The last stage of our experiment consists of testing ANN, SNN-I and SNN-II with different types of adversarial inputs. The adversarial inputs are passed through the forward function of the networks and compared against the true labels to compute the corresponding adversarial test accuracy and loss. 
We have performed four different sets of comparisons, as described in the "Testing" section in Fig. \ref{fig:workflow}. We have trained two separate networks with the same architecture, but different initialization for each of ANN, SNN-I and SNN-II. They are labelled as $M_{ANN}$, $M_{ANN^x}$; $M_{SNN-I}$, $M_{SNN-I^x}$ and $M_{SNN-II}$, $M_{SNN-II^x}$. 
\begin{enumerate}
    \item Whitebox: In this scenario, each of the target models ($M_{ANN}$, $M_{SNN-I}$ and $M_{SNN-II}$) is attacked by the adversarial input generated from their respective target network. 
    \item SNN-I-crafted Blackbox: In this set of comparison, all of the target models are attacked by inputs crafted from a single SNN-I model $M_{SNN-I^x}$.
    \item SNN-II-crafted Blackbox: In this case, all of the target models are attacked by inputs crafted from a single SNN-II model $M_{SNN-II^x}$.
    \item ANN-crafted Blackbox: Here the common source model for all three targets ($M_{ANN}$, $M_{SNN-I}$, $M_{SNN-II}$) is $M_{ANN^x}$. 
     
\end{enumerate}
The target and source models for these comparisons are summarized in table \ref{table:attack_comparisons}.

\begin{table*}[!t]
\caption{Adversarial attack comparisons: ANN, SNN-I and SNN-II}
\label{table:attack_comparisons}
\begin{center}
    \begin{tabular}{| c | c | c | c | c | c | c |}
    \hline
     &\multicolumn{2}{|c|}{\textbf{ANN}}&\multicolumn{2}{|c|}{\textbf{SNN-I}}&\multicolumn{2}{|c|}{\textbf{SNN-II}} \\ \hline
     & Target & Source & Target & Source & Target & Source \\ \hline
     \textbf{Whitebox}& $M_{ANN}$ & $M_{ANN}$ & $M_{SNN-I}$ & $M_{SNN-I}$ & $M_{SNN-II}$ & $M_{SNN-II}$ \\ \hline
     \textbf{SNN-I-crafted Blackbox}& $M_{ANN}$ & $M_{SNN-I^x}$ & $M_{SNN-I}$ & $M_{SNN-I^x}$ & $M_{SNN-II}$ & $M_{SNN-I^x}$ \\ \hline
     \textbf{SNN-II-crafted Blackbox}& $M_{ANN}$ & $M_{SNN-II^x}$ & $M_{SNN-I}$ & $M_{SNN-II^x}$ & $M_{SNN-II}$ & $M_{SNN-II^x}$ \\ \hline
     \textbf{ANN-crafted Blackbox}& $M_{ANN}$ & $M_{ANN^x}$ & $M_{SNN-I}$ & $M_{ANN^x}$ & $M_{SNN-II}$ & $M_{ANN^x}$ \\ \hline
    \hline
    \end{tabular}
\end{center}
\end{table*}

\section{Results}\label{sec:results}

\subsection{ANN versus SNN adversarial robustness}\label{sec:results_ann_vs_snn}
\begin{figure*}[!t]
\centerline{\includegraphics{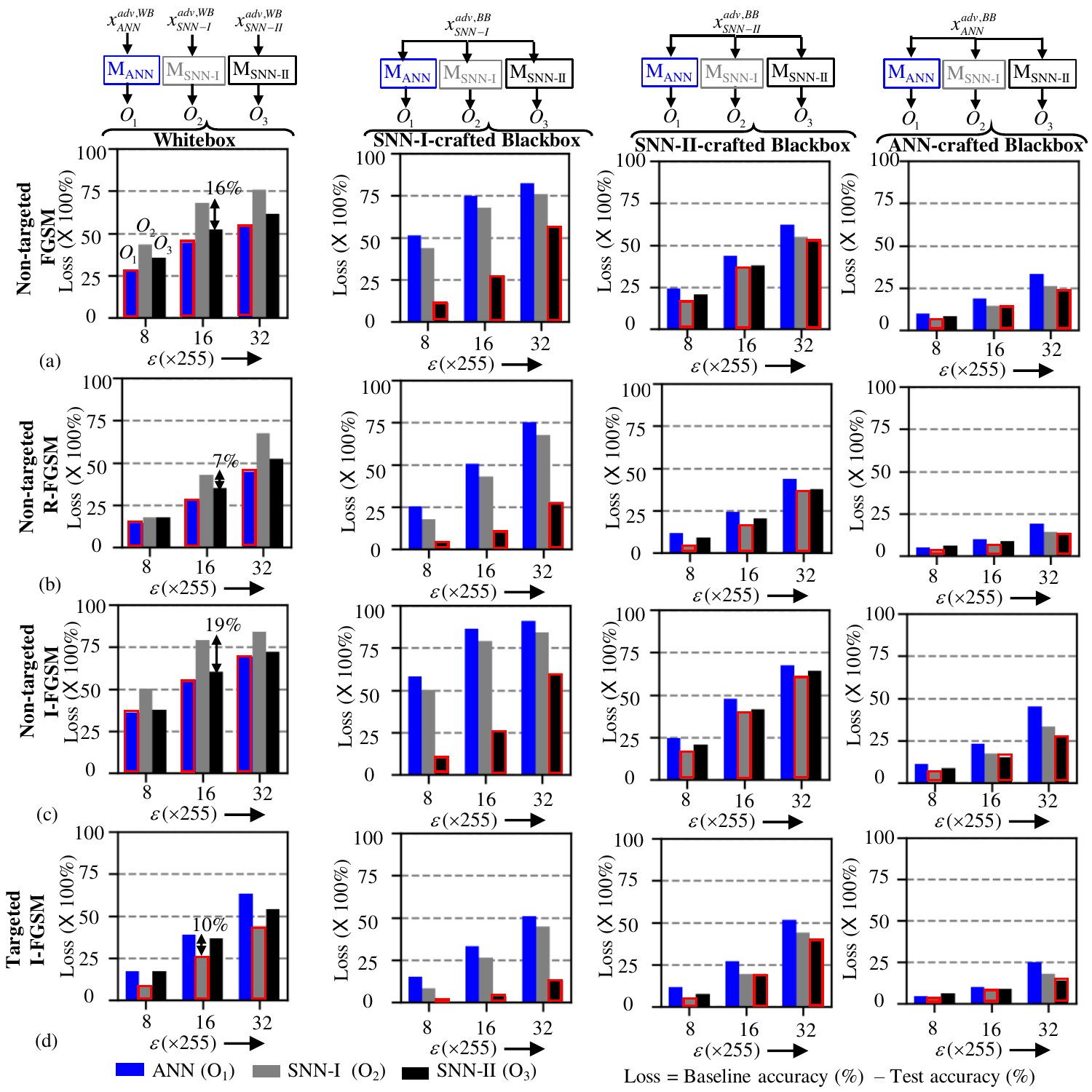}}
\caption{A comparison of the loss in accuracy under adversarial attack of ANN (blue bar, $O_1$), SNN-I (Grey bar, $O_2$) and SNN-II (Black bar, $O_3$) for four different types of attacks: (a) Non-targeted FGSM, (b) Non-targeted R-FGSM, (c) Non-targeted I-FGSM (2-step) and (d) Targeted I-FGSM (2-step). In each of these cases, the amount of perturbation $\epsilon$ has been varied from 8/255 to 32/255. Column 1 presents whitebox attack results. Columns 2-4 present blackbox attack results, wherein the networks are attacked by a common set of adversarial inputs (SNN-I crafted blackbox adversary for column 2, SNN-II crafted blackbox adversary for column 3 and ANN-crafted blackbox adversary for column 4). Note, the method detailed earlier in Fig. \ref{fig:workflow} is used to generate SNN-crafted adversary from SNN-I and SNN-II. The loss for each network is calculated as the difference between the baseline accuracy (clean input test) and test accuracy (adversarial input test). Smaller loss in accuracy implies more adversarial robustness. Minimum loss values are highlighted with red rectangles in each subplot.  
}
\label{fig:attack_types}
\end{figure*}

First, we compared the ANN and SNN behavior under whitebox scenario (column 1 of Fig. \ref{fig:attack_types}), where each target network is attacked by adversarial inputs created using the target's parameters (ANN is attacked by ANN-crafted, SNN-I by SNN-I crafted and so on). It is evident from column 1 in Fig. \ref{fig:attack_types} that ANN faces smaller degradation in accuracy compared to SNN-I and SNN-II against different kinds of whitebox attacks (Non-targeted {FGSM, R-FGSM, I-FGSM} scenarios) for varying $\epsilon$ ranges. For instance, in FGSM whitebox attack (Fig. \ref{fig:attack_types}(a) column 1),  when $\epsilon$ = 8/255, $O_1$ (ANN loss), $O_2$ (SNN-I loss) and $O_3$ (SNN-II loss) correspond to 27.8\%, 43.5\% and 35.5\%, respectively. However, we observed that with a targeted attack (specifically, targeted I-FGSM shown in Fig. \ref{fig:attack_types} (d) Column 1), SNN-I and SNN-II losses are lower than ANN. 

Next, we compared the robustness of the models against blackbox attacks. It is worth mentioning that ANN and SNN models differ in terms of network dynamics and adversarial input generation mechanism. Hence, to have a fair comparison in blackbox scenario, we  used a common source model (separately trained and different from the target model) to generate the adversaries and subsequently, attacked ANN, SNN-I and SNN-II with it, as illustrated in Fig. \ref{fig:attack_types}. The source model used to generate adversarial example is SNN-I, SNN-II and ANN in column 2, 3 and 4, respectively. Interestingly, in contrary to the whitebox results, we observe that SNNs turn out to be more robust in blackbox setting. Even for the ANN-crafted blackbox attack case (column 4), where all three networks yield smaller accuracy degradation, SNNs show lower loss than ANN. This points to the fact that spike-based computing with temporal dynamics has some intrinsic resistance, compared to standard rate-based ANN dynamics. We conjecture that the stochasticity in the temporal dynamics that is inherited with spike computing might be contributing to this adversarial resistance. Another noteworthy observation here is that the accuracy loss observed with SNN-I/-II-crafted blackbox attacks across all the models is significantly higher than ANN-crafted blackbox attacks. This implies that adversarial inputs generated with temporal spike-based SNNs cast stronger attacks than rate-based ANNs. Revisiting the whitebox attack results, we can surmise that the stronger attack created from SNN models is the cause of higher accuracy degradation for spiking models in that case.

In summary, we can deduce the following from the above results with regard to ANN vs. SNN adversarial effects: 1) SNNs cast stronger adversarial attacks than ANNs, leading to more accuracy loss for SNN, compared to ANN, in whitebox scenario. 2) SNNs are more robust in blackbox setting than ANNs due to the inherent stochastic temporal dynamics.

\subsection{Dependence on SNN training method: SNN-I versus SNN-II}\label{sec:results_snn1_vs_snn2}

\begin{figure}[!tb]
\centerline{\includegraphics{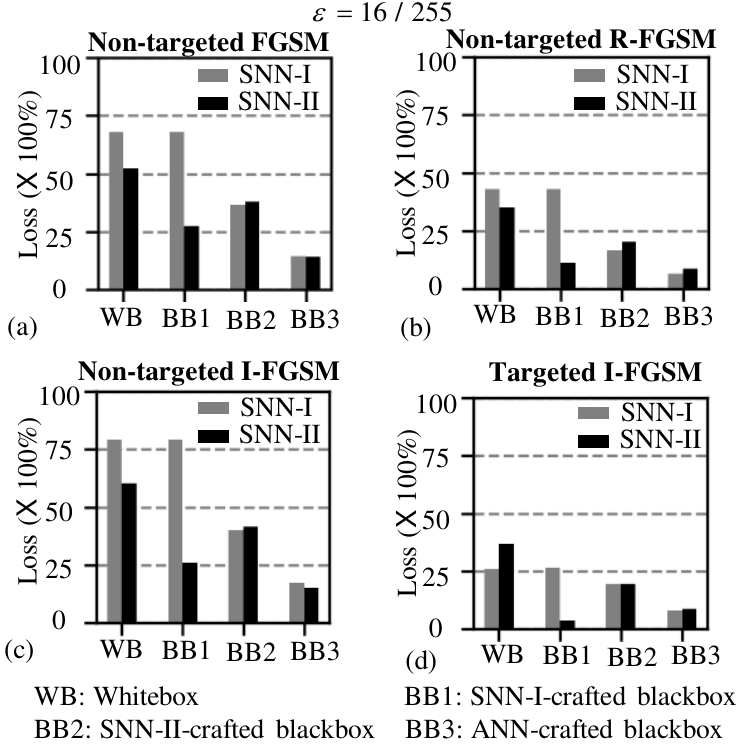}}
\caption{A comparison of SNN-I and SNN-II loss for four types of adversarial inputs: (a) Non-targeted FGSM, (b) Non-targeted R-FGSM, (c) Non-targeted I-FGSM (2-step) and (d) Targeted I-FGSM (2-step) at $\epsilon = 16/255$. Each subplot contains comparison results under whitebox (WB: SNN-I is attacked by SNN-I-crafted, SNN-II is attacked by SNN-II-crafted input), SNN-I-crafted blackbox (BB1: both SNN-I and SNN-II are attacked by a separately trained SNN-I), SNN-II-crafted blackbox (BB2: both SNN-I and SNN-II are attacked by a separately trained SNN-II) and ANN-crafted-blackbox (BB3: both SNN-I and SNN-II are attacked by ANN-crafted input) scenario.}
\label{fig:snn1_snn2}
\end{figure}

 Now, we analyse the dependence of the training mechanism (used to create the SNN) on its adversarial resistance. In order to obtain a clearer picture, we compared the adversarial resistance of SNN-I (ANN-to-SNN conversion) and SNN-II (direct spike based training) against different kinds of attacks, described as follows (Fig. \ref{fig:snn1_snn2}):
\begin{itemize}
    \item Whitebox (WB): Here, SNN-I (SNN-II) is attacked by adversarial input produced from the same SNN-I (SNN-II), respectively. During this WB attacks in Fig. \ref{fig:snn1_snn2} (a)-(d), SNN-I undergoes more loss compared to SNN-II for all types of attacks (except Targeted I-FGSM).
    \item{SNN-I-crafted Blackbox (BB1): Here, the source model for adversary generation is a separately trained SNN-I (different from the target SNN-I). This common adversary is used to attack both SNN-I and SNN-II. For all types of adversaries, SNN-II exhibits significant robustness compared to SNN-I.}
    \item{SNN-II-crafted Blackbox (BB2): The common set of adversary, in this case, is obtained from a separately trained SNN-II. Previously, for BB1 (SNN-I-crafted attack), we observed that the accuracy loss of SNN-I was significantly higher than SNN-II. Hence, in this case, one might expect that SNN-II target models will yield very high accuracy loss compared to SNN-I targets (since the attack is crafted from SNN-II). However, the amount of accuracy degradation for SNN-II is still comparable to that of SNN-I. This suggests that networks trained with spike-based training have more adversarial resistance that conversion-based models. We speculate that the deterministic nature of conversion based models (converting ReLU values to IF functionality) does not entirely inherit the stochasticity in temporal dynamics causing adversarial susceptibility for SNN-I models.}
    \item{ANN-crafted Blackbox (BB3): In this scenario, the common adversarial source model is an ANN. Here the loss values for SNN-I and SNN-II lies within a range of 1\% for all attack types and are much lower than the BB1, BB2 scenarios. This further corroborates the fact that SNNs craft stronger attacks than ANNs.}
\end{itemize}
Note, Fig. \ref{fig:snn1_snn2} basically contains the adversarial test results for SNN-I and SNN-II presented in the previous figure (Fig. \ref{fig:attack_types}) at a fixed $\epsilon$ value ($\epsilon = 16/255$).

\begin{figure}[!tb]
\centerline{\includegraphics{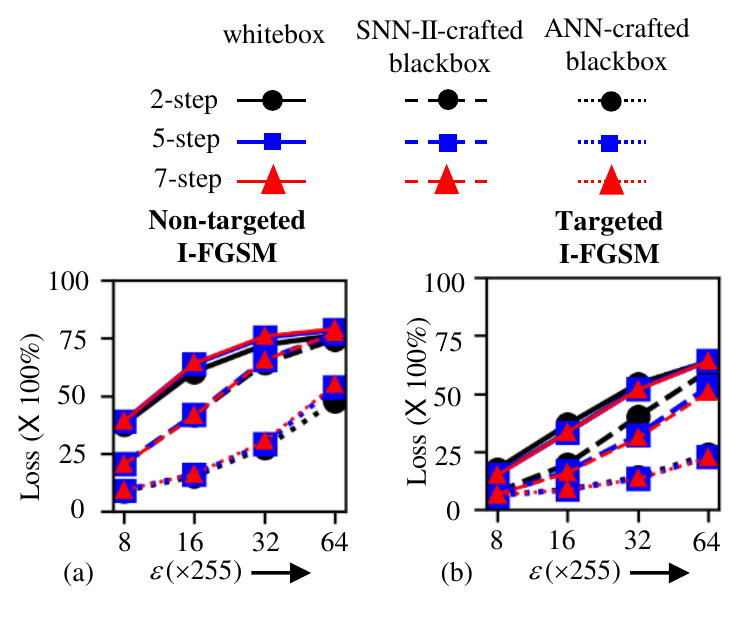}}
\caption{Comparison of the loss in SNN-II under I-FGSM (a) Non-targeted and (b) Targeted attacks for number of steps = 2, 5, 7, when perturbation $\epsilon$ is varied from 8/255 to 64/255. In each plot, solid, dashed and dotted lines indicate attack by SNN-II-crafted whitebox input, SNN-II-crafted blackbox input and ANN-crafted blackbox input, respectively. The magnitude of loss is mostly independent of the number of steps used in I-FGSM and yields minimum value when the network is attacked by ANN-crafted input for all cases.}
\label{fig:ifgsm_stepsize}
\end{figure}

\subsection{Dependence on I-FGSM iteration steps for SNN-II}\label{sec:results_ifgsm_step}
In order to investigate the effect of the number of iteration steps used in I-FGSM on SNN, we have plotted the loss of SNN-II for Non-targeted and Targeted I-FGSM attacks with two-step (black), five-step (blue) and seven-step (red) iterative methods in Fig. \ref{fig:ifgsm_stepsize}(a) and (b). Note, iterative attacks are stronger than single-step attacks (such as, FGSM or R-FGSM). I-FGSM input for the target model ($M_{SNN-II}$) has been created from three different source networks: $M_{SNN-II}$ (whitebox: solid lines), $M_{SNN-II^x}$ (SNN-II-crafted-blackbox: dashed lines) and $M_{ANN^x}$ (ANN-crafted blackbox: dotted lines). We note that the step-number variation has little effect on the performance of the models in all cases. In addition, as observed earlier, SNN-crafted blackbox inputs cause stronger attacks compared to ANN-crafted blackbox. To analyse this further, we compared the slope of the lines in Targeted I-FGSM attack (Fig. \ref{fig:ifgsm_stepsize}(b)). Amount of loss rises from 6\% to only 22\% for ANN-crafted blackbox, as we increase $\epsilon$ (the amount of adversary) from 8/255 to 64/255, whereas loss due to SNN-crafted blackbox attack (dashed line) undergoes a staggering increase from 6\% to 50\%. This establishes the effectiveness of SNN-based adversaries for casting stronger attacks. 

\section{Conclusion}\label{sec:conclusion}
In this paper, we analyzed the role of bio-plausible Spiking Neural Networks in the domain of adversarial attacks. As an initial work in this field, we have addressed some unexplored issues like introducing simplified method for crafting adversarial sample from spiking networks and dependence of robustness on SNN training mechanism. In addition, our quantitative comparison for variety of adversarial attacks presents a comprehensive picture of the performance of these networks under different attack scenarios. Finally, the key findings and recommendations from our analysis are: 
\begin{itemize}
\item{ SNNs craft stronger attack than ANNs in both whitebox and blackbox setting.}
\item{While SNNs undergo higher accuracy degradation than ANNs in whitebox scenarios, they yield significantly higher resistance than ANNs for blackbox attacks. The temporal dynamics and inherent stochasticity in SNNs might be responsible for such behavior. Further work is required to understand the role of temporal dynamics for adversarial resistance.}
\item{SNNs trained with spike based backpropagation are more robust than SNNs obtained from conversion rules against both whitebox and blackbox attacks. This further ascertains the role of stochasticity (preserved in the spike-based backpropagation  mechanism) to strengthen adversarial resistance.}

\end{itemize}


\section*{Acknowledgment}
The work was supported in part by, Center for Brain-inspired Computing (C-BRIC), a DARPA sponsored JUMP center, Semiconductor Research Corporation, National Science Foundation, Intel Corporation, the DoD Vannevar Bush Fellowship and U.S. Army Research Laboratory.


\vspace{12pt}


\begin{thebibliography}{00}
\bibitem{b1} A. Krizhevsky, I. Sutskever and G. E. Hinton, Imagenet classification with deep convolutional neural networks. In Advances in neural information processing systems, pp. 1097-1105, 2012.

\bibitem{hinton} Geoffrey Hinton, Li Deng, Dong Yu, George Dahl, Abdel rahman Mohamed, Navdeep Jaitly, Andrew Senior, Vincent Vanhoucke, Patrick Nguyen, Tara Sainath, and Brian Kingsbury. Deep neural networks for acoustic modeling in speech recognition. Signal Processing Magazine, 2012.

\bibitem{szegedy} C. Szegedy, W. Zaremba, I. Sutskever, J. Bruna, D. Erhan, I. Goodfellow, R. Fergus, Intriguing properties of neural networks, arXiv preprint arXiv:1312.6199, 2014.

\bibitem{goodfellow} Ian J. Goodfellow, Jonathon Shlens, and Christian Szegedy. Explaining and harnessing adversarial examples. CoRR, abs/1412.6572, 2014. URL http://arxiv.org/abs/1412.6572.

\bibitem{papernot} Nicolas Papernot, Patrick Drew McDaniel, Ian J. Goodfellow, Somesh Jha, Z. Berkay Celik, and Ananthram Swami. Practical black-box attacks against deep learning systems using adversarial examples. CoRR, abs/1602.02697, 2016a. URL http://arxiv.org/abs/1602.02697.

\bibitem{kurakin2} A. Kurakin, I. J. Goodfellow, S. Bengio, Adversarial examples in the physical worls, workshop track - ICLR 2017.

\bibitem{sharif} Mahmood Sharif, Sruti Bhagavatula, Lujo Bauer, and Michael K. Reiter. Accessorize to a crime:
Real and stealthy attacks on state-of-the-art face recognition. In Edgar R. Weippl, Stefan Katzenbeisser, Christopher Kruegel, Andrew C. Myers, and Shai Halevi, editors, Proceedings of the 2016 ACM SIGSAC Conference on Computer and Communications Security, Vienna, Austria, October 24-28, 2016, pages 1528–1540. ACM, 2016.


\bibitem{ensemble} F. Tramer et. al., Ensemble adversarial training: attacks and defenses, ICLR 2018.

\bibitem{exL} P. Panda, K. Roy, Implicit Generative Modeling of Random Noise during Training for Adversarial Robustness, arXiv:https://arxiv.org/abs/1807.02188, 2018.

\bibitem{kurakin} A. Kurakin, I. J. Goodfellow, S. Bengio, Adversarial machine learning at scale, ICLR 2017.

\bibitem{snn} Maass, W. (1995). “On the computational power of noisy spiking neurons,” in Proceedings of the 8th International Conference on Neural Information Processing Systems (Denver, CO; Cambridge, MA: MIT Press), 211–217.

\bibitem{abhronil} A. Sengupta, Y. Ye, R. Wang, C. Liu, K.Roy, "Going Deeper in Spiking Neural Networks: VGG and Residual architectures", Front. Neurosci. 13:95. doi: 10.3389/fnins.2019.00095

\bibitem{diehl}P. U. Diehl, D. Neil, J. Binas, M. Cook, S.-C. Liu, and M. Pfeiﬀer, “Fast-classifying, high-accuracy spiking deep networks through weight and threshold balancing,” in Neural Networks (IJCNN), 2015 International Joint Conference on. IEEE, 2015, pp. 1–8. 

\bibitem{direct_train1} C. Lee, P. Panda, G. Srinivasan, K. Roy, Training Deep Spiking Convolutional Neural Networks With STDP-Based Unsupervised Pre-training Followed by Supervised Fine-Tuning, Front Neurosci. 2018; 12: 435.
\bibitem{direct_train} Chankyu Lee, Syed Shakib Sarwar, Kaushik Roy, Enabling Spike-based Backpropagation in State-of-the-art Deep Neural Network Architectures, 
arXiv:1903.06379, 2019.

\bibitem{pytorch} A. Pazske et. al., Automatic differentiation in PyTorch, NIPS 2017 Workshop Autodiff.

\end{thebibliography}
\end{document}